\def\year{2022}\relax
\documentclass[letterpaper]{article} 
\usepackage{aaai22}  
\usepackage{times}  
\usepackage{helvet}  
\usepackage{courier}  
\usepackage[hyphens]{url}  
\usepackage{graphicx} 
\usepackage{natbib}  
\usepackage{caption} 
\DeclareCaptionStyle{ruled}{labelfont=normalfont,labelsep=colon,strut=off} 
\usepackage{algorithm}

\usepackage{newfloat}
\usepackage{listings}
\floatstyle{ruled}
\newfloat{listing}{tb}{lst}{}
\floatname{listing}{Listing}
\title{Discourse-level Relation Extraction via Graph Pooling}
\author {
    I-Hung Hsu \textsuperscript{\rm 1},
    Xiao Guo \textsuperscript{\rm 2},
    Premkumar Natarajan \textsuperscript{\rm 1},
    Nanyun Peng \textsuperscript{\rm 1, \rm 3}
}
\affiliations {
    \textsuperscript{\rm 1} Information Sciences Institute, University of Southern California\\
    \textsuperscript{\rm 2} Department of Computer Science and Engineering, Michigan State University\\
    \textsuperscript{\rm 3} Department of Computer Science, University of California, Los Angeles\\
    \texttt{\{ihunghsu, pnataraj\} @isi.edu \\ guoxia11@msu.edu \\ violetpeng@cs.ucla.edu}
}

\usepackage{times}
\usepackage{latexsym}

\usepackage{graphicx}
\newcommand{\mypar}[1]{\noindent\textbf{#1}}
\usepackage{amsfonts}
\usepackage{booktabs}
\usepackage{tabularx}
\usepackage{tabulary}
\usepackage{amsmath,amsthm}
\usepackage{makecell}
\usepackage{multirow}
\usepackage{algorithm}
\usepackage[noend]{algorithmic}
\usepackage{soul}

\usepackage{color}
\definecolor{mygreen}{RGB}{89, 169, 90}
\definecolor{myblue_2}{RGB}{56, 89, 157}
\definecolor{myblue}{RGB}{0, 173, 229}
\definecolor{myorange}{RGB}{247, 144, 61}
\definecolor{DarkRed}{RGB}{130,25,0}

\usepackage{microtype}
\usepackage{todonotes}
\usepackage{caption}

\begin{document}

\maketitle
\begin{abstract}
The ability to capture complex linguistic structures and long-term dependencies among words in the passage is essential for discourse-level relation extraction (DRE) tasks.
Graph neural networks (GNNs), one of the methods to encode dependency graphs, have been shown effective in prior works for DRE.
However, relatively little attention has been paid to \textit{receptive fields} of GNNs, which can be crucial for cases with extremely long text that requires discourse understanding.
In this work, we leverage the idea of \textit{graph pooling} and propose to use pooling-unpooling framework on DRE tasks. The pooling branch reduces the graph size and enables the GNNs to obtain larger \textit{receptive fields} within fewer layers; the unpooling branch restores the pooled graph to its original resolution so that representations for entity mention can be extracted.
We propose Clause Matching (CM), a novel linguistically inspired graph pooling method for NLP tasks.
Experiments on two DRE datasets demonstrate that our models significantly improve over baselines when modeling long-term dependencies is required, which shows the effectiveness of the pooling-unpooling framework and our CM pooling method.
\end{abstract}
\section{Introduction}
\label{Introduction}
\begin{figure}[t!]
\centering
\includegraphics[width=0.93\linewidth]{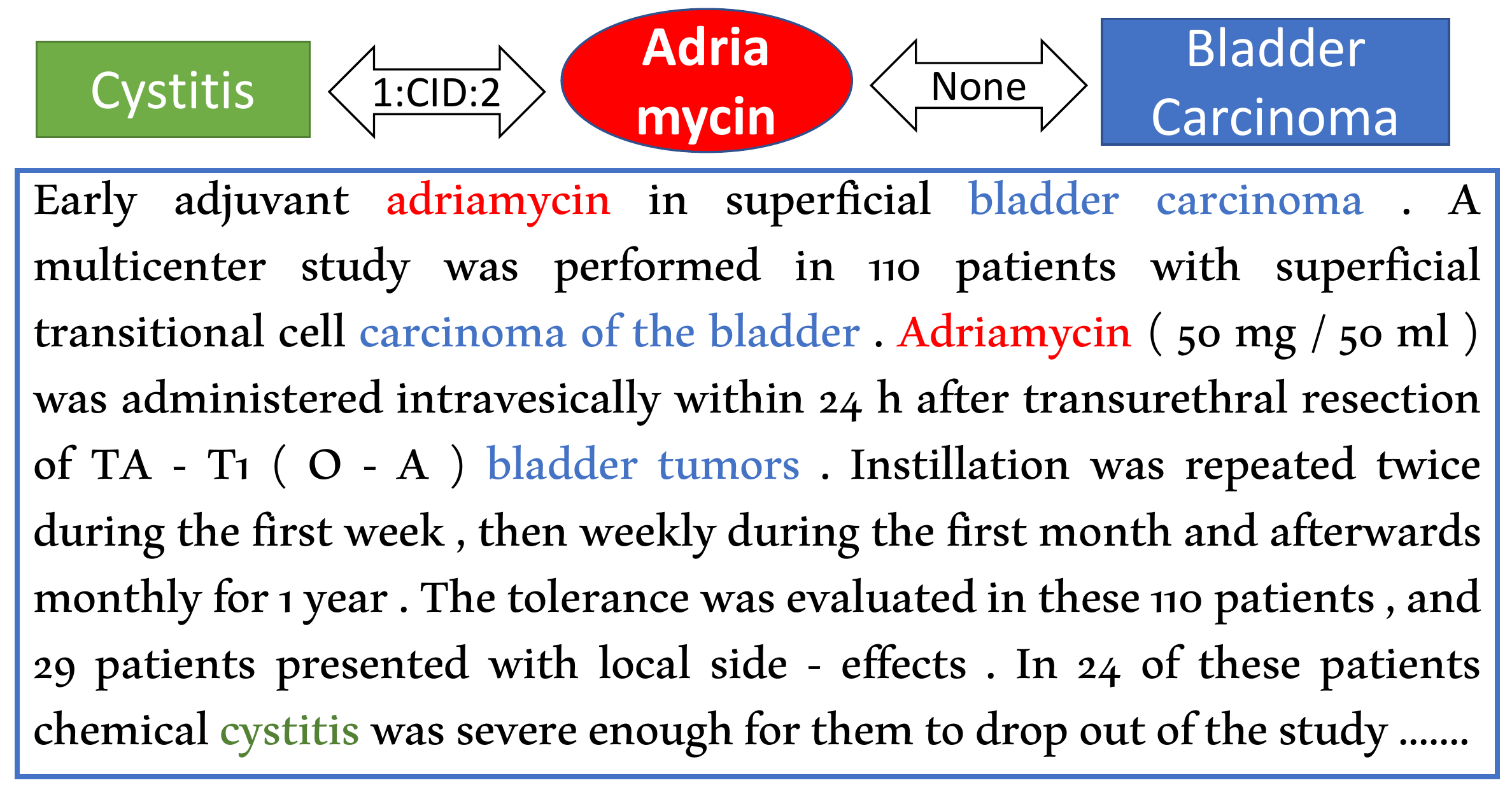}
\caption{An example of DRE from the CDR dataset. The oval represents chemical \textbf{\textit{Adriamycin}}, and rectangles represent two diseases: \textbf{\textit{Bladder Carcinoma}} and \textbf{\textit{Cystitis}}. Reading the whole passage over six sentences is essential to the understanding the relation between \textbf{\textit{Adriamycin}} and \textbf{\textit{Cystitis}}.}
\label{intro_fig}
\end{figure}

Relation extraction, the task to extract the relation between entities in the text, is an important step for automatic knowledge base construction \cite{DBLP:conf/doceng/Al-ZaidyG17, dong2014knowledge}.
While earlier works in relation extraction focus on relations within a single sentence \cite{DBLP:conf/emnlp/XuFHZ15,DBLP:conf/acl/MiwaB16}, recent works place more emphasis on identifying the relations that appear at discourse level \cite{DBLP:conf/naacl/JiaWP19,DBLP:conf/aaai/XuCZ21}, which is more practical yet challenging to models. Most notably, models need to consider larger amount of information and capture the \textit{long-term dependencies} between words to catch relations between entities spanning several sentences. 
For example, in Fig.~\ref{intro_fig}, the relation between \textit{Cystitis} and \textit{Adrimycin} can only be understood by considering the whole paragraph.

\begin{figure*}[t!]
    \centering
    \includegraphics[trim=0cm 0cm 0cm 0cm, clip, width=0.84\linewidth]{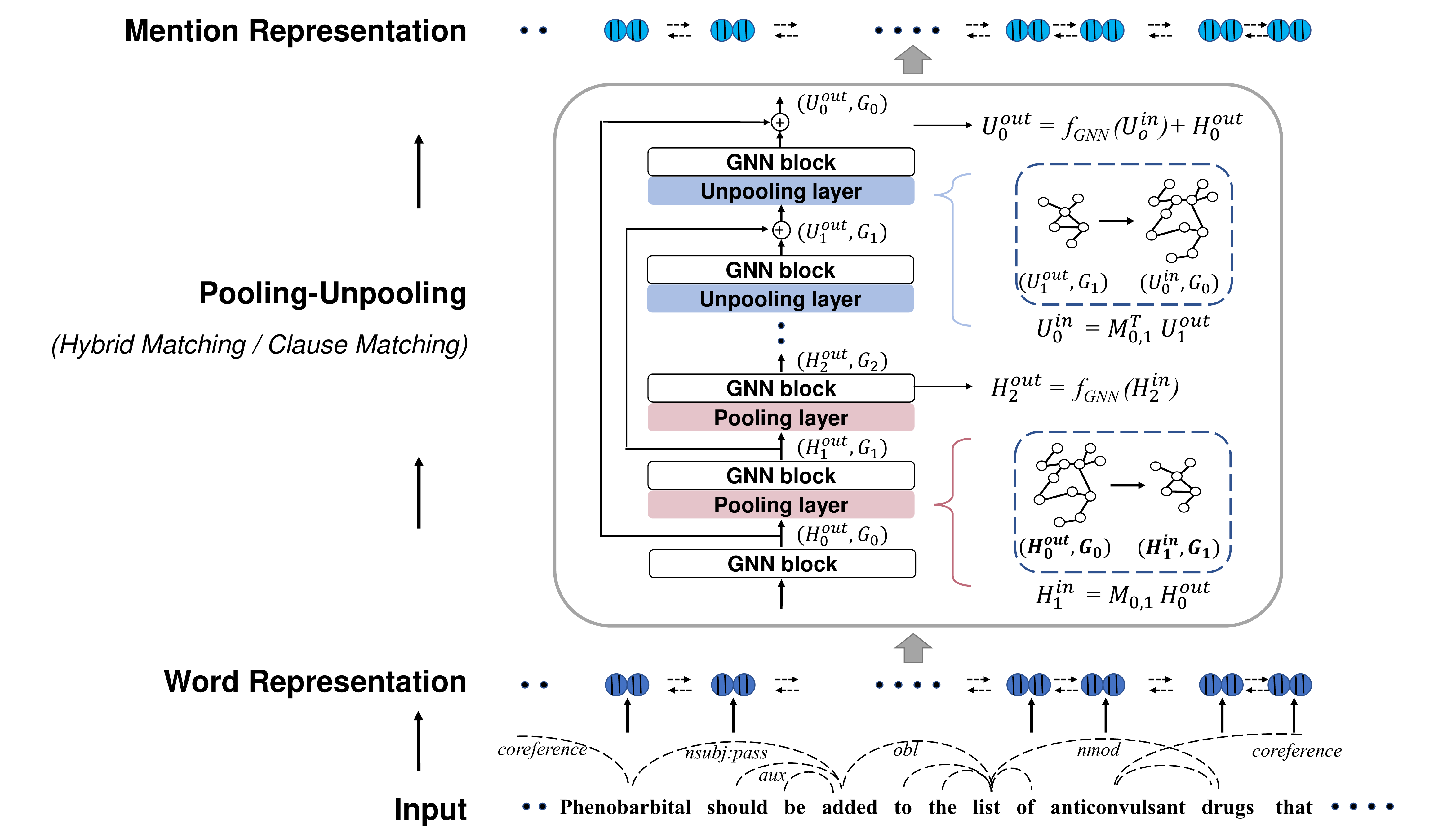}
	\caption{Our proposed pooling-unpooling framework: Input words are first encoded into vectors and then fed into pooling-unpooling layers to learn the structural dependencies within the input tokens. GNN block can be any model that can encode graphs and update node embeddings. Pooling layers merges nodes in the original graph into supernodes, forming smaller size hypergraphs. Unpooling layers map the information from the coarser graph back to finer-grained graph. 
	In the figure, $G_i$ represents hypergraphs at the $i$-th level. $H_i^{in}$ and $U_i^{in}$ denote the converted embeddings after the pooling and unpooling layers in $i$-th hypergraph. Both of them can be calculated using the matching matrix $M_{l,l+1}$. We use $H_i^{out}$ and $U_i^{out}$ to denote the refined representations after GNN blocks.}
	\label{Architecture}
\end{figure*}

To capture such long-term dependencies, previous works incorporate dependency trees to capture syntactic clues in a non-local manner \cite{DBLP:conf/emnlp/Zhang0M18,nary}, which help the model to effectively explore a broader context with structure. Graph neural networks (GNNs) have been widely applied in this case~\cite{nary, DBLP:conf/emnlp/SongZWG18, DBLP:conf/acl/SahuCMA19}. However, the \textit{receptive fields}~\cite{DBLP:conf/nips/LuoLUZ16} of GNNs, which measure the information range that a node in a graph can access, are less discussed. In theory, it is essential for GNNs to obtain a larger receptive field for learning representations to better capture extremely long-term dependencies. It is non-trivial for prior methods \cite{nary, DBLP:conf/emnlp/SongZWG18, DBLP:conf/acl/SahuCMA19} to achieve this requirement and usually requires additional efforts, such as properly stacking a deep network and without being potential saturation like oversmoothing~\cite{DBLP:conf/aaai/LiHW18}.

To alleviate this effort and facilitate the representation learning for entity mentions in discourse-level relation extraction (DRE) tasks, in this work, we leverage the idea of \textit{graph pooling} and propose to use a pooling-unpooling framework (as depicted in Figure~\ref{Architecture}). In the pooling branch, we use graph pooling to convert the input graph to a series of more compact hypergraphs by merging structurally similar or related nodes. The node representation learning on the hypergraph thus aggregates a larger neighborhood of features compared to the learning on the original graph, increasing the size of the receptive field for each node. Then, we use unpooling layers to restore the global information in hypergraphs back to the original graph so that embeddings for each entity mention can be extracted. With such a pooling-unpooling mechanism, each graph node can obtain richer features.

We use graph convolutional network (GCN) \cite{DBLP:conf/iclr/KipfW17}, one of the most commonly used GNNs, to instantiate our proposed pooling-unpooling framework. Additionally, we explore two graph pooling strategies -- Hybrid Matching and Clause Matching to justify our method's generalizability. Hybrid Matching (HM)~\cite{DBLP:conf/icwsm/LiangG021} merges nodes based on the structural similarity and has got success in learning large-scale graph. However, HM ignores the edge type information when merges nodes, which could be helpful to group nodes in DRE tasks. For example, nodes linked with \textit{``nmod''} dependency edge can often be merged when considering its linguistic meaning in text\footnote{\textit{``nmod''} represents it is a nominal dependents of another noun}. 
Thus, we propose Clause Matching (CM) to leverage the information regarding dependent arcs' type to merge nodes. 
Comparing to HM, which is a general graph pooling algorithm that merges nodes by considering the overall graph structure, CM is designed from the linguistic perspective and specifically focuses on the dependent relations between nodes.
Despite differences, our proposed method with either pooling method achieves improvements over baseline on two DRE datasets -- CDR~\cite{CDR} and cross-sentence $n$-ary~\cite{nary}. We further carry out a comprehensive analysis to understand the proposed framework to verify the pooling-unpooling framework's effectiveness to handle cases that requiring extreme long-term dependencies.

\section{Method}
\label{sec:method}
In this section, we introduce our proposed pooling-unpooling framework for DRE, which is illustrated in Figure~\ref{Architecture}. The input to the system is a document graph that contains nodes as tokens and edges as the tokens' various dependencies (Sec.~\ref{sec:input_graph}).
The node embeddings before pooling-unpooling layers are word representations, which can be either static word representations like GloVe or contextualized embedding like BERT~\cite{BERT}. The node representations are then fed into our pooling-unpooling framework, which consists of a series of GNN blocks, pooling layers, and unpooling layers. The framework updates node representations so that they can better capture long-term dependencies. 
Specifically, the pooling layer deterministically converts a graph into a more compact \textit{hypergraph} using \textit{matching matrices} generated by graph pooling methods (Sec.~\ref{Graph_pooling}).
The unpooling layer performs a reverse operation to the pooling layer, restoring finer-grain graphs from the hypergraphs to the original graph (Sec.~\ref{ssec:unpooling}).
Between each pooling and unpooling layers, GNN blocks are employed to update node embeddings so that the information within nodes can be exchanged.
After our pooling-unpooling layers, we can then extract the entity mention representations and perform DRE tasks (Sec.~\ref{ApplyMGCN}).
\subsection{Input Graph}
\label{sec:input_graph}
We adapt document graph~\cite{DBLP:conf/eacl/QuirkP17} to represent intra- and inter-sentential dependencies in texts to help models effectively explore a broader context of structure.
Document graph consists of nodes representing words and edges representing various dependencies among words. Typically, two words can be linked if they (1) are adjacent, (2) have dependency arcs, or (3) share discourse relations, such as coreference or being roots of sequential sentences.
We use document graphs to represent input texts and apply our model to them for leveraging syntactic and discourse clues.
\subsection{Graph Pooling}
\label{Graph_pooling}
Given the input graph $G$, our next step is to use \textit{graph pooling} to iteratively coarsens $G$ into a smaller but structurally similar graph $G'$.
Graph pooling methods use various methods to discover nodes that can be grouped~\cite{DBLP:conf/nips/YingY0RHL18,DBLP:conf/icml/LeeLK19}, and then, nodes that are matched together will be merged into a \textit{supernode}. In this paper, we explore two different graph pooling methods -- \textit{Hybrid Matching} and our proposed \textit{Clause Matching}.

\begin{figure}[!t]
    \centering 
    \includegraphics[width=1.0\columnwidth]{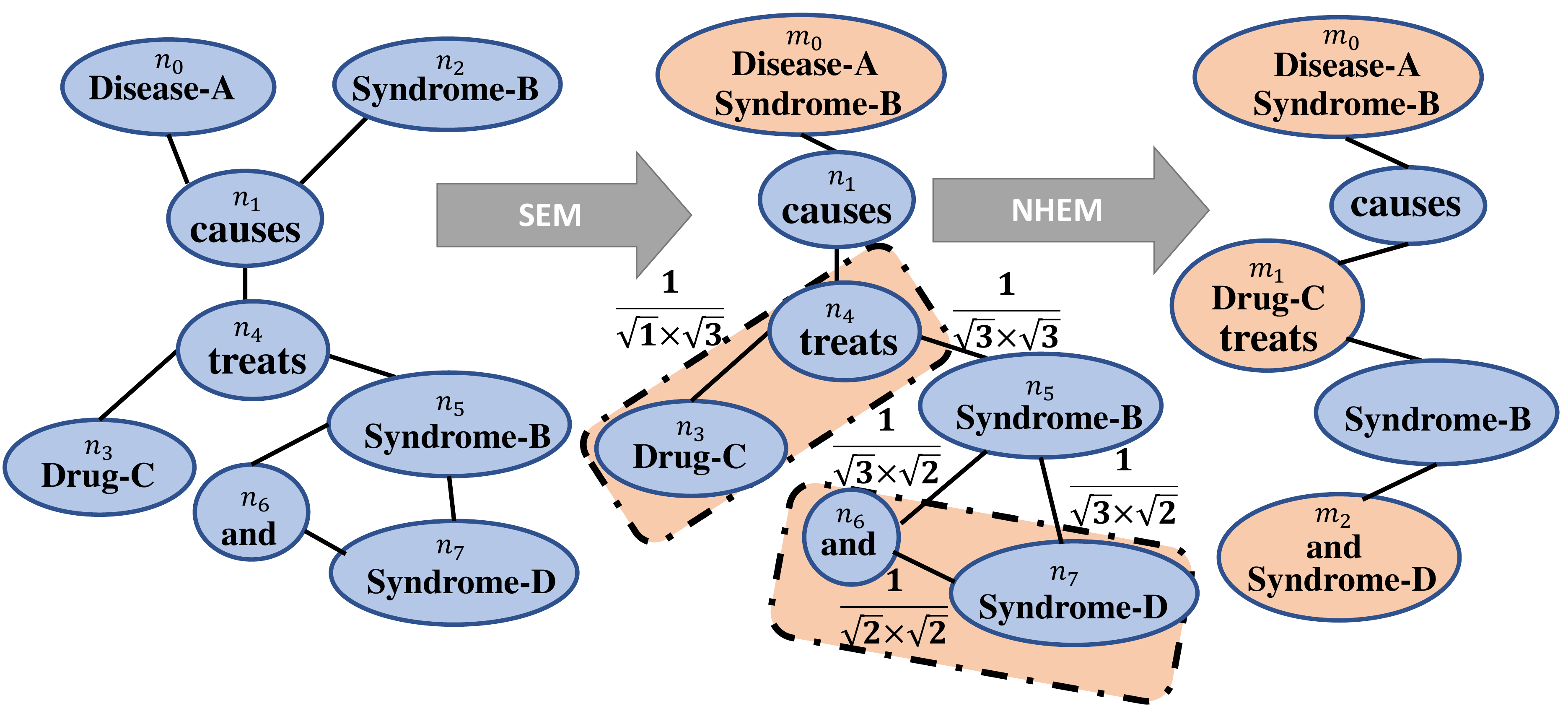} 
    \caption{The illustration of Hybrid Matching (HM) for the example, \textit{``Disease-A causes Syndrome-B. Drug-C treats Syndrome-B and Syndrome-D''} with target entities \textit{``Disease-A''}, \textit{``Drug-C''} and \textit{``Syndrome-D''}. After performing HM, the distances from \textit{``Drug-C''} to \textit{``Disease-A''}, and from \textit{``Drug-C''} to \textit{``Syndrome-D''} decrease by 1.} 
    \label{fig:hybrid_matching} 
\end{figure}

\paragraph{Hybrid Matching.}
Hybrid Matching (HM), proposed by \citet{DBLP:conf/icwsm/LiangG021}, has been shown effective for encoding large-scale graphs. It performs node matching based on the connectivity between nodes and consists of two steps: (1) \textit{structural equivalence matching} (\textsc{SEM}), and (2) \textit{normalized heavy edge matching} (\textsc{NHEM}). 

\textsc{SEM} merges two nodes that share the exact same neighbor. In the Figure~\ref{fig:hybrid_matching} example, the node $\textit{$n_0$}$ and the $\textit{$n_2$}$ are considered as structural equivalence, since they share the exact same neighbor, the node $\textit{$n_1$}$. 

On the other hand, \textsc{NHEM} uses the graph's adjacency matrix $A$ to perform node matching. Each node will be matched with its neighbor that has the largest \textit{normalized edge weight}, under the constraint that supernodes cannot be merged again. 
\textit{Normalized edge weight}, $N(u, v)$, for the edge between the node $u$ and the node $v$ is:
\begin{equation}
\small
N(u, v)=\frac{A_{uv}}{\sqrt{D(u)\times D(v)}},\label{eq:NHEM}
\end{equation}
where $D(u)$ is the degree of the node $u$.
The adjacency matrix $A$ of the original input document graph (i.e, $A^0$) has cell value $A^0_{uv}$ being either 1 or 0, indicating if a connection from node $u$ to $v$ exists or not in the document graph. For the adjacency matrix for pooled graphs, it can be calculated using Equation~\ref{matching_eq_0}, which will be discussed in the later paragraph. When performing NHEM, we visit nodes by ascending order according to the nodes' degree following \citet{DBLP:conf/icwsm/LiangG021}. We explain the process using the example in Figure~\ref{fig:hybrid_matching}. After computing the normalized edge weight, we first visit the node $\textit{$n_3$}$ (degree equals 1), and merge it with its only neighbor $\textit{$n_4$}$, forming the supernode $\textit{$m_1$}$. Then, we visit the node $\textit{$n_6$}$ and merge it with $\textit{$n_7$}$, which has the largest normalized edge weight with $\textit{$n_6$}$. Since supernodes cannot be merged again, $\textit{$n_1$}$ and $\textit{$n_5$}$ remain. In the Figure~\ref{fig:hybrid_matching} example, after executing HM, we can observe that the distance between targeted entities decreases.

\paragraph{Clause Matching.}
The edge attribute could be important to match nodes in the document graph. For example, \textit{``a red apple''} can be split into three nodes in the document graph, although they are essentially one noun phrase. The dependency tree shows that \textit{``a''} is the determiner (det) of \textit{``apple''}, and \textit{``red''} is an adjectival modifier (amod) of \textit{``apple''}. Such edge information could be useful in the pooling operation, but has been ignored by many graph pooling method like HM. Inspired by this observation, we propose to merge tokens considering the dependency arcs' type and introduce Clause Matching (CM).

\begin{algorithm}[b]
\footnotesize	
  \caption{Clause Matching}\label{CM algorithm}
  \begin{algorithmic}[1]
    \REQUIRE{$G=(V,E)$, $E\subseteq \{e=(v_i, v_j)|(v_i, v_j)\in V^2\}$.}
    \REQUIRE{Edge type function $f_T$: $f_T(e)$= the edge type of $e$.}
    \REQUIRE{Mergeable edge type set $T$.}
    \STATE{Sort $V$ by the number of neighbors in ascending order.}
    \STATE{$C=\emptyset$} \textit{\# The set collects nodes that have merged others.}
    \FOR{$v_i\in V$}
        \IF{$v_i \notin C$}
            \FOR{$v_j\in \{v_j|v_j\in V;(v_j, v_i)\in E\}$}
                \IF{$f_T(e=(v_j,v_i)) \in T$}
                    \STATE $V=V\setminus\{v_i\}$ \textit{\# Merge $v_i$ to $v_j$}
                    \FOR{$v_k\in \{v_k|v_k\in V;(v_k, v_i)\in E\}$} 
                        \STATE \textit{ \# Move edges in $v_i$ to $v_j$}
                        \STATE $E=E\setminus\{(v_k, v_i)\}$ 
                        \STATE $E=E\cup\{(v_k, v_j)\}$ 
                    \ENDFOR
                    \STATE $C=C\cup\{v_j\}$
                    \STATE \textbf{break}
                \ENDIF
            \ENDFOR
        \ENDIF
    \ENDFOR
    \RETURN{Pooled graph $G=(V, E)$}
  \end{algorithmic}
\end{algorithm}

Specifically, we reference the Universal Dependencies project ~\cite{DBLP:conf/lrec/NivreMGHMPSTZ20}, which provide detailed definitions of each dependency relation, to design the CM algorithm by applying our domain knowledge.
We first classify all dependency relations into two categories -- \textit{core arguments} and \textit{others}. \textit{Core arguments} link predicates with their core dependents, as a clause should at least consist of a predicate with its core dependents. CM will merge tokens that are connected by dependency relations in the \textit{others} categories, e.g., ``det'' and ``amod'' are not \textit{core arguments} but \textit{others.}. In the example \textit{``a red apple''}, by merging the two edges, \textit{``a red apple''} becomes one node that represents the noun phrase. Since we do not merge nodes linked with \textit{core arguments}, the basis of a clause will be retained. As a result, CM simplifies the graph while maintaining the core components of a clause.

The details of CM are presented in Algorithm~\ref{CM algorithm}~\footnote{We define the set $T$ as \{all dependency edges and coreference edges\} $\setminus$ \{``nsubj'', ``nsubj:pass'', ``dobj'', ``iobj'', ``csubj'', ``csubj:pass'', ``ccomp'', ``xcomp''\}.}. CM share similarities with HM in two points: (1) we visit nodes by ascending order according to the node degree (line 1); (2) supernodes cannot be merged again in each round of CM (line 4). Being different from HM, we decide whether the visited node can be merged with its dependent head based on the edge type (line 6-7)~\footnote{When moving edges from children nodes to supernodes, we do not include self-loop causing by merging.}.

\begin{figure}[!t]
    \centering 
    \includegraphics[width=1.0\columnwidth]{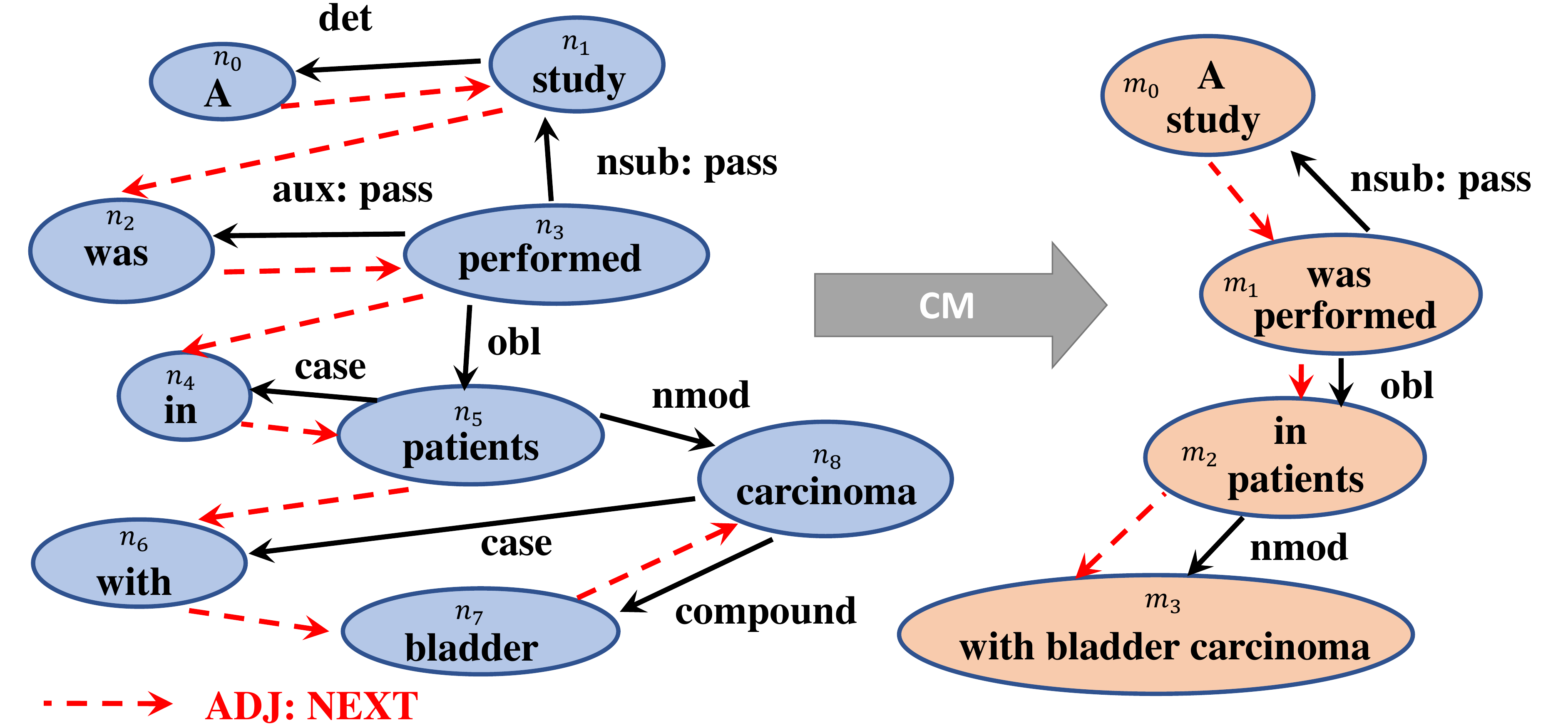} 
    \caption{The visualization of Clause Matching (CM), given the input \textit{``A study was performed in patients with bladder carcinoma''}. ``ADJ:NEXT'' indicates the adjacency edges in the document graph. After executing CM once, the graph size reduces largely yet still maintain the core structure of the original sentence.} 
    \label{fig:clause_matching} 
\end{figure}

We illustrate CM with an example in Figure~\ref{fig:clause_matching}. In Figure~\ref{fig:clause_matching}, we will visit nodes $\textit{$n_0$}$, $\textit{$n_2$}$, $\textit{$n_4$}$, ..., $\textit{$n_5$}$ in order, based on their node degree. When visiting $\textit{$n_0$}$, CM matches $\textit{$n_0$}$ with its dependent head $\textit{$n_1$}$ because the dependent arc between them belongs to the \textit{others} types. Similarly, $\textit{$n_2$}$ is merged with $\textit{$n_3$}$ and $\textit{$n_4$}$ is matched with $\textit{$n_5$}$, forming $\textit{$m_1$}$ and $\textit{$m_2$}$, respectively. However, $\textit{$m_2$}$ cannot be further combined with $\textit{$m_1$}$ because $\textit{$m_2$}$ is already a supernode in the current round of CM. 
Moreover, $\textit{$m_0$}$ cannot be merged with $\textit{$m_1$}$ even if we perform CM again because the dependency arc ``nsubj:pass'' belongs to \textit{core arguments}. In contrast, if we perform CM again to the hypergraph, $\textit{$m_2$}$ will be merged with $\textit{$m_1$}$. More visualizations for CM can be found in the Appendix~\ref{appendix:visual_cm}.

\paragraph{Pooling Layer.}
\label{Pooling_Operation}
The main function that pooling layers perform is to map the node representations from input graph $G$ to its coarsened hypergraph $G'$. To achieve the mapping, we rely on the \textit{matching matrix} that we get during graph pooling.

Each graph pooling process generates a coarsened hypergraph. Performing matching $L$ times produces $L$ hypergraphs with increasing coarsening levels, denoted as $G_0, G_1, ..., G_L$, where $G_0$ is the initial document graph. We use the matching matrix $M_{l-1,l}$ to mathematically represent the merging process from level $l-1$ to level $l$~\footnote{The matching matrix can be obtained together with the graph pooling algorithm. We omit the matching matrix calculation step in our Algorithm 1 block for simplicity.}.

$M_{l-1,l} \in \mathbb{R}^{n \times m}$ converts $G_{l-1} \in \mathbb{R}^{n \times n}$  to $G_l \in \mathbb{R}^{m \times m}$ with $n \geq m$. Each cell $m_{ij}$ in $M_{l-1,l}$ is:
\begin{equation}
\small
    m_{ij}= 
\begin{cases}
    1, &\small{\text{if node $i$ is matched into supernode $j$.}}\\
    0, &\text{otherwise.}
\end{cases}
\end{equation}

With $M_{l-1,l}$ constructed, we can compute the adjacency matrix of level $l$, $A^l$, based on $A^{l-1}$\footnote{$A^0$ is the adjacency matrix of input document graph.}:
\begin{equation}
\label{matching_eq_0}
\small
A^l=M^T_{l-1,l}A^{l-1}M_{l-1,l},
\end{equation}
and perform representation mapping to get the initial node embeddings for the next level's graph $G_{l}$:
\begin{equation}
\label{matching_eq_1}
\small
H_l^{in} = M_{l-1,l}H^{out}_{l-1},
\end{equation}
where $H^{out}_{l-1}$ and $H^{in}_{l}$ represent the output node embedding of $G_{l-1}$ and the input embedding of $G_{l}$. 
\subsection{Graph Unpooling} 
\label{ssec:unpooling}
After layers of graph pooling, the model can encode node information that are originally far in the input graph. The unpooling branch is then designed to restore the information to the original resolution. Specifically, the unpooling layers use the matching matrices, which are generated during graph pooling, to perform reverse operations, including generating larger graphs and mapping the embeddings from coarsened graphs to unpooled graphs. We denote the unpooling layer's function on $l$-th graph embedding as:
\begin{equation}
\small
U^{in}_{l-1} = M^T_{l-1,l}U^{out}_{l}.
\label{Eq_unpooling}
\end{equation}
To reinforce the learning of graph nodes, each unpooling layer is followed by a GNN block to finetune node representations in the graph. The function is written as $\widetilde{U^{out}_{l-1}} = f_{\text{GNN}}(U^{in}_{l-1})$, as shown in Figure~\ref{Architecture}\footnote{We use GCN~\cite{DBLP:conf/iclr/KipfW17} as our GNN in this work.}.
Additionally, to combine the information at different scales, we add residual connections~\cite{DBLP:conf/cvpr/HeZRS16} to refine our representation, i.e. $U^{out}_{l-1} = \widetilde{U^{out}_{l-1}} + H^{out}_{l-1}$. 

\subsection{Apply to Relation Extraction}
\label{ApplyMGCN}
The token embeddings obtained after our pooling-unpooling framework encode comprehensive features. In this paper, we use these features for DRE tasks with minimal design in order to test the quality of the extracted mention representation.

Given an entity mention span, we simply extract the corresponding mention representation by using max-pooling on the tokens within the given span. This setting is commonly used in previous works~\cite{DBLP:conf/acl/GuoZL19}. 
When prediction relations between entity mentions, we then concatenate all the targeting entity mention representations with additional max-pooled sentence embedding, and then feed them into linear layers to get the mention-pair score. If the DRE dataset only provides relation labels at the entity level, we follow \citet{DBLP:conf/naacl/JiaWP19} to use the LogSumExp to aggregate information from multiple mention-pairs:
\begin{equation}
\label{eq:logsumexp}
\small
score(E_1, E_2) = log\sum_{e_1 \in E_1, e_2 \in E_2}exp(g(e_1, e_2)),
\end{equation}
where $score(E_1, E_2)$ is the final logit for entity pair ($E_1$, $E_2$), and $g(e_1,e_2)$ is the mention-pair score for the given mention tuple $e_1$ and $e_2$. 

\begin{table}[t!]
\centering
\small
\begin{tabular}{@{}l@{ }|@{ }l@{ }|@{ }c@{ }c@{ }c@{ }c@{}}\toprule
\multicolumn{2}{@{ }l@{ }|}{\multirow{2}{*}{\textbf{Model}}}                                                               & \multicolumn{2}{c@{}}{\textbf{Detection}} & \multicolumn{2}{c@{}}{\textbf{Classification}} \\ \cline{3-6} 
\multicolumn{2}{@{}l@{}|}{}                                                                                     & Ternary           & Binary          & Ternary             & Binary            \\ \hline
\multicolumn{2}{@{}l@{}|}{Graph LSTM \cite{nary}}                                                                           & 80.7           & 76.7          & -                & -                \\ 
\multicolumn{2}{@{}l@{}|}{AGGCN$^*$ (Guo et al. \shortcite{DBLP:conf/acl/GuoZL19})}                                                                                & 76.7           & 79.0            & 67.5             & 67.9             \\ \hline
\multirow{2}{*}{\begin{tabular}[c]{@{}l@{}}GloVe\\ Embedding\end{tabular}} & Pooling (HM) (ours) & 82.2           & \textbf{80.8}          & \textbf{76.2}             & \textbf{75.1}             \\ 
                                                                           & Pooling (CM) (ours) & \textbf{82.5}           & \textbf{80.8}          & 75.4             & 73.4             \\ \hline \hline
\multirow{4}{*}{SciBERT}                                                   & Finetune & 83.2           & 82.7          & 78.5             & 76.8             \\ 
                                                                           & GCN       & 83.7           & 83.1          & 78.9             & 77.2             \\ \cline{2-6}
                                                                           & Pooling (HM) (ours) & 84.4           & \textbf{83.4}          & \textbf{79.6}             & \textbf{78.0}             \\
                                                                           & Pooling (CM) (ours) & \textbf{84.5}           & \textbf{83.4}          & 79.1             & \textbf{78.0}             \\ 
\bottomrule
\end{tabular}
\caption{Results of five-fold cross-validation on the $n$-ary dataset in average accuracy(\%). $^{*}$ indicates our re-run via their release code with hyperparameter tuning.} 
\label{tab:main_res}
\end{table}

\section{Experiments}
\label{results}
To instantiate our proposed pooling-unpooling framework, we choose to use GCN \cite{DBLP:conf/iclr/KipfW17} as the GNN block to conduct our experiments on two DRE datasets. First, we perform experiments on the Cross-sentence $n$-ary dataset, a DRE dataset focusing on mention-level relations. This is to test the quality of our extracted mention representations. Then, we experiment on the Chemical-Disease Reactions dataset, a DRE dataset studying entity-level relations. Experiments on CDR can justify whether our method can capture entity-level relations even with minimal design.
Dataset statistics and the best hyper-parameters are in Appendix~\ref{data_detail}\&\ref{hyper_detail}.

\subsection{Cross-sentence $n$-ary Dataset}
\label{N_ary_main_res}
\mypar{Data and Task Settings:}
The cross-sentence $n$-ary dataset ($n$-ary)~\cite{nary} contains drug-gene-mutation \textit{ternary} and drug-mutation \textit{binary} relations annotated via distant supervision \cite{DBLP:conf/acl/MintzBSJ09}. Data is categorized with 5 classes: “resistance or non-response”, “sensitivity”, “response”, “resistance”, and “None”. Following prior works, the \textit{detection task} treats all relation labels except ``None'' as True. In the experiments, we replace target entity mentions with dummy tokens to prevent the classifier from simply memorizing the entity name. This is a standard practice in distant supervision RE~\cite{DBLP:conf/naacl/JiaWP19}\footnote{We call this setting \textit{Entity Anonymity} in contrast to the setting where all tokens are exposed, denoted as \textit{Entity Identity} setup. Prior works are inconsistent w.r.t. this experimental detail. Specifically, \citet{nary} conducted experiments under \textit{Entity Anonymity} setup, while \citet{DBLP:conf/emnlp/SongZWG18} and \citet{DBLP:conf/acl/GuoZL19} reported results under \textit{Entity Identity} setup. For fair comparisons, we report our results under \textit{Entity Identity} in the Appendix \ref{appendix:entity_id_nary}.}. The document graphs of the dataset are provided in the original release. All other data setup follows \citet{DBLP:conf/emnlp/SongZWG18}.
Our evaluation follows previous measurements -- average accuracy of 5-fold cross-validation.

\paragraph{Compared baselines:} (1) \textbf{Graph LSTM} \cite{nary}, which modifies the LSTM structures so that they can encode graphical inputs. (2) \textbf{AGGCN} \cite{DBLP:conf/acl/GuoZL19}, which is a GCN-based model with adjacency matrices that operate in the GCN being adjusted by attention mechanism. (3) \textbf{Finetune} pretrained word representations. Considering that the $n$-ary dataset is in biomedical domain, we use SciBERT~\cite{DBLP:conf/emnlp/BeltagyLC19}. (4) \textbf{GCN} with pretrained word representations, which add additional GCN layers after pretrained word representation layers.

\begin{table}[t!]
\centering
\small
\setlength{\tabcolsep}{3pt}
\begin{tabular}{l|l|c}\toprule
\multicolumn{2}{l|}{\textbf{Model}} 
          & \textbf{Test F1}     \\\hline 
\multirow{7}{*}{\textbf{Static Emb}} & GCCN \cite{DBLP:conf/acl/SahuCMA19}                  & 62.3            \\
& EOG  (Christopoulou et al. \shortcite{DBLP:conf/emnlp/ChristopoulouMA19})                  & 63.6  \\
& LSR \cite{DBLP:conf/acl/NanGSL20}       & 64.8   \\
& EoGANE (Tran et al. \shortcite{DBLP:conf/emnlp/TranNN20})                            & \textbf{66.1}      \\ \cline{2-3}
& GCN (our implementation)   & 62.7    \\ 
& Pooling (HM) (ours)$^\dagger$      & 64.7     \\   
& Pooling (CM) (ours)$^\dagger$    & \underline{65.6}    \\ \midrule 
\multirow{4}{*}{\textbf{BERT Large}} & SSAN$_\text{Biaffine}$  \cite{DBLP:conf/aaai/XuWLZM21}   & 65.3    \\ \cline{2-3}
& GCN (our implementation)       & 65.1    \\ 
& Pooling (HM) (ours)    & \underline{66.1}      \\
& Pooling (CM) (ours)$^\dagger$     & \textbf{66.3}   \\ \midrule
\multirow{6}{*}{\textbf{SciBERT}} & SSAN$_\text{Biaffine}$  \cite{DBLP:conf/aaai/XuWLZM21}       & \underline{68.7}    \\
& ATLOP \cite{DBLP:conf/aaai/Zhou0M021}      & \textbf{69.4} \\
& ATLOP$^*$ \cite{DBLP:conf/aaai/Zhou0M021}   & \underline{69.1} \\ \cline{2-3} 
& GCN (our implementation)         & 66.8    \\ 
& Pooling (HM) (ours)$^\dagger$   & 68.0      \\
& Pooling (CM) (ours)$^\dagger$   & 68.2  \\ \bottomrule
\end{tabular}
\caption{Results measured in F1(\%) on CDR dataset. The underlined numbers representing the second best scores and the bold numbers are the best for each fair comparison. $^{*}$ indicates our re-run using their released code with 3 runs and select the best run. $^\dagger$ indicates the model outperforms its corresponding GCN baseline significantly (with p-value $<$ 0.05 per McNemar's test).}
\label{tab:cdr_res}
\end{table}

\begin{figure*}[!t]
    \centering 
    \includegraphics[width=1.0\linewidth]{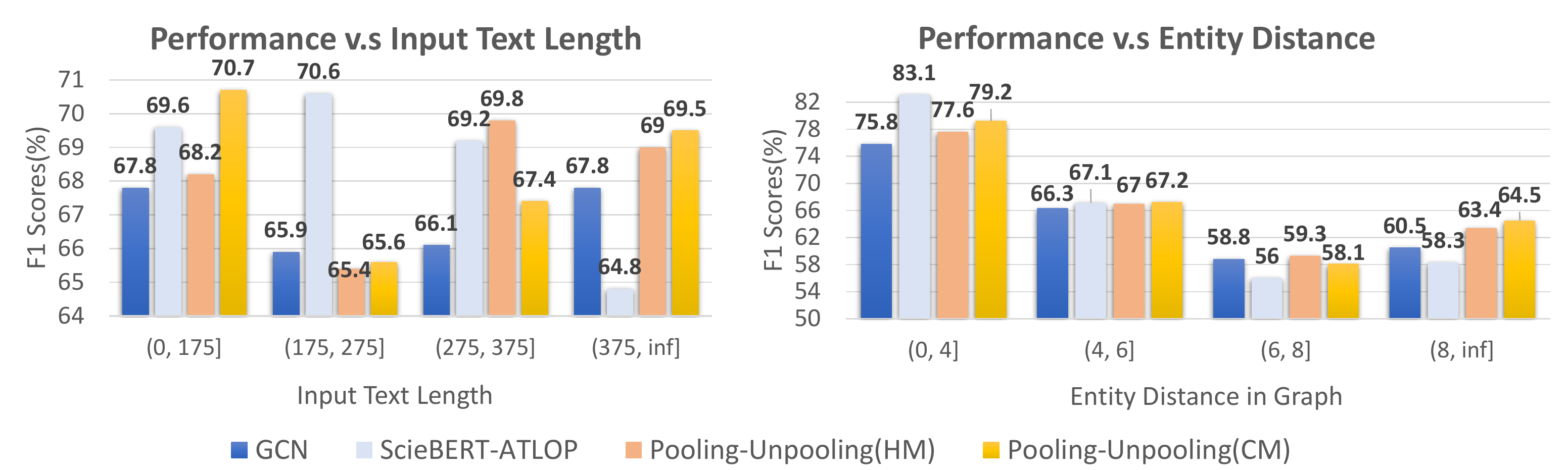} 
    \caption{Models' performance against input text length and the entity distance in graph.} 
    \label{fig:cdr_by_length} 
\end{figure*}

\paragraph{Results on $n$-ary dataset:} Table~\ref{tab:main_res} shows the comparison on the $n$-ary dataset. We report results of our method, Pooling-Unpooling(HM) and Pooling-Unpooling(CM), representing our method with Hybrid Matching and with Clause Matching, respectively. 
To fairly compare our method with baselines without contextualized word representation (Graph LSTM and AGGCN), we also report our result with GloVe embedding \cite{DBLP:conf/emnlp/PenningtonSM14}.

Comparing to prior works without contextualized word representation, the results show that both of our methods surpass the baseline by at least $1.8\%$ average accuracy. 
Under the situation of using pretrained SciBERT, our results consistently outperform the GCN baseline, which indicates that the pooling-unpooling framework is helpful to learn a better mention representation for DRE tasks. Additionally, the result of GCN consistently outperforms SciBERT finetuning. This provides empirical evidence that the learning of structure dependencies is helpful to perform DRE tasks even though a strong pretrained word representation is provided.

Lastly, we compare CM with HM. In contrast of HM that takes all edges into consideration, CM only uses the dependency arc types to group nodes. In the original document graph in $n$-ary dataset, there are some missing dependency edges, which prevents CM from merging nodes. HM, which can address the situation because of its consideration on overall graph structure, shows its robustness under this condition. Although getting into the situation, CM still achieves competitive results with HM. This highlights the importance of considering edge type information in the graph pooling stage.

\subsection{Chemical-Disease Reactions Dataset}
\mypar{Data and Task Settings:}
The Chemical-disease reactions dataset (CDR)~\cite{CDR} is a DRE task that focuses on entity level relations. No document graph is provided in the original CDR datset, so we generate its document graph using Stanford CoreNLP~\cite{DBLP:conf/acl/ManningSBFBM14}. We follow \citet{DBLP:conf/emnlp/ChristopoulouMA19}\footnote{We also follow them to use (1) GENIA Sentence Splitter for sentence splitting (2) GENIA Tagger~\cite{tsuruoka2005developing} for tokenization (3) PudMed pre-trained word embeddings \cite{DBLP:conf/bionlp/ChiuCKP16} for static embedings.} to train the model in two steps. First, we train our model using standard training set and record the best hyperparameter when the model reaches optimal on the development set. Then, the model is re-trained using the union of the training and development data for final test. 

\paragraph{Compared baselines:} We consider prior literature on the CDR dataset \cite{DBLP:conf/acl/SahuCMA19, DBLP:conf/emnlp/ChristopoulouMA19, DBLP:conf/acl/NanGSL20, DBLP:conf/emnlp/TranNN20, DBLP:conf/aaai/XuWLZM21, DBLP:conf/aaai/Zhou0M021}. Among these works, SSAN~\cite{DBLP:conf/aaai/XuWLZM21} studies on the most similar goals to us, i.e, toward achieving better entity mention representation for DRE tasks. Hence, we conduct experiments like them, which studies different word representations, BERT Large \cite{BERT} and SciBERT~\cite{DBLP:conf/emnlp/BeltagyLC19}. 
To fairly compare with the baselines, we also include our results with static embeddings.~\footnote{Our models with static embeddings contains additional LSTM layers after the static word embedding. This is a fair comparison setting with other baselines using static embeddings.}
Additionally, we add a \textbf{GCN} baseline to verify the effectiveness of our proposed pooling-unpooling mechanism.

\paragraph{Results on CDR dataset:} 
The result is shown in Table~\ref{tab:cdr_res}. 
First, we compare our method with prior works. Our models can surpass most of the previous works even with minimal design on aggregating information from mention representations to entity embeddings.
Then, we compare our method with SSAN. Although SSAN finishes with a slightly better test F1 score with SciBERT (0.5 F1 difference), our models outperform them under the condition of using BERT Large with a 1.2 F1 margin.
Third, from the table, we observe that CM is able to outperform HM on all three situation. This, again, verifies the effectiveness of CM.

Lastly, we study the effectiveness of the proposed pooling-unpooling framework. Comparing our pooling-unpooling models with our GCN baselines, the pooling-unpooling models achieve significant improvement regardless of which pretrained word representation being used. 

\subsection{Studies}
In this section, we perform studies to better understand our method. Specifically, we provide analysis on two questions: (1) whether the pooling-unpooling mechanism helps the learning of long-term dependencies; (2) how sensitive is our model regarding the quality of graph pooling?

\paragraph{The learning for long-term dependencies:} 
To answer the first question, we analyze our model by examining (1) the model's performance against the input text length, (2) the model's performance against the entity distance in the input document graph\footnote{We compute the entity distance in the graph by first calculating the minimum distance between each targeting mention-pair in the graph, then taking the maximum distance among all mention-pairs. Such a design is to estimate the largest effort for models to capture the current relations.}. 
We conduct experiments on the CDR dataset using models with SciBERT and report the results that evaluated on the test set.
We compare our models with our GCN baseline and SciBERT-ATLOP~\cite{DBLP:conf/aaai/Zhou0M021}, and present the comparison in Figure~\ref{fig:cdr_by_length}.
From the figure, we observe that models with pooling-unpooling can perform better for cases that need to resolve long dependencies (instances with larger entity distances or instances with longer inputs).

\paragraph{Models using different graph pooling algorithm:} 
We conduct experiments with models using different graph pooling methods to test how our method is affected by the graph pooling algorithm. We consider three additional pooling methods: (1) \textbf{Pooling-Unpooling (Random Pooling)}: models with random pooling strategy. During graph pooling, each edge has the probability of 0.5 to be used for merging nodes; (2) \textbf{Pooling-Unpooling (Aggressive HM)}: for each graph pooling layer, i.e, from $G_i$ to $G_{i+1}$, we perform HM twice. That is, the size of the output graph after one aggressive HM is much smaller than that after one standard HM; (3) \textbf{Pooling-Unpooling (Aggressive CM)}: similar to aggressive HM, we perform twice CM for each graph pooling layer. We run experiments on the CDR dataset with SciBERT embedding and report the performance on the development set in Table~\ref{table:various_pool}.

From the table, we observe that: (1) The pooling quality is important. Comparing random pooling with other variations, we see that the result is almost the same as pure GCN. (2) Aggressive HM can achieve even better results than HM, yet, aggressive CM yields worse results than CM. We hypothesize that this is because CM has already been more aggressive than HM. Based on our statistics, the averaged graph size for the development size in CDR is 258.39 tokens. After one HM pooling, the averaged graph size reduces to 150.45, while the average graph size after CM decreases to 90.34. Hence, performing aggressive CM may lead to information loss because of being too aggressive, yet aggressive HM can make the receptive fields being enlarged more efficiently. To sum up, the pooling quality matters in our framework, and how to get the best graph pooling approach is still an area that needs more exploration.

\begin{table}[!t]
\centering
\small
\begin{tabular}{l|c}\toprule
\textbf{Model}             & \textbf{Dev F1} \\
\hline
GCN                                & 68.4   \\
Pooling-Unpooling (HM)             & 69.2   \\
Pooling-Unpooling (CM)             & 69.5   \\
\hline
Pooling-Unpooling (Random Pooling) & 68.6   \\
Pooling-Unpooling (Aggressive HM)  & 69.4   \\
Pooling-Unpooling (Aggressive CM)  & 68.9  \\
\bottomrule
\end{tabular}
\caption{Models' performance with various graph pooling methods. The experiments are measured in F1(\%) on the development set of the CDR dataset.} 
\label{table:various_pool} 
\end{table}

\section{Related work}
\label{sec:related}
\mypar{Discourse-level Relation Extraction.}
There have been growing interests for studying relation extraction beyond single sentence. 
Many of these works~\cite{DBLP:conf/acl/GuoZL19, DBLP:conf/aaai/GuptaRSR19, DBLP:conf/ijcai/GuoN0C20} introduce intra- and inter-sentential structure dependencies, such as dependency trees, to help capture representations for target mentions. Various GNNs are proposed and applied to encode this structural information~\cite{nary, DBLP:conf/emnlp/SongZWG18, DBLP:conf/aaai/XuWLZM21}. Yet, their model designs mostly neglect the efforts of their models to capture long-term dependencies when the input graph growing to discourse level.

Another thread of works on DRE is to identify relations of entity pairs (i.e., relations between entities rather than a single pair of mentions) from the entire document~\cite{DBLP:conf/naacl/JiaWP19,DBLP:conf/emnlp/ChristopoulouMA19,DBLP:conf/emnlp/ZengXCL20,DBLP:conf/aaai/Zhou0M021}. \citet{DBLP:conf/naacl/JiaWP19} propose a multiscale learning paradigm, which learns final relation representations by gradually constructing mention-level, entity-level representation.
\citet{DBLP:conf/emnlp/ChristopoulouMA19} use rules to build edge representation from extracted nodes and study on using edge representations to make final prediction.
\citet{DBLP:conf/emnlp/ZengXCL20} first extract mention representations using a simple encoder, and then design mention-level and entity-level graphs to capture the interplay among mentions and entities.
These methods mostly put emphasis on learning the interaction between the mention representations and entity embeddings, while our main goal is to learn better mention representations that capture long-term dependencies.

\paragraph{Graph Pooling in NLP.}
Graph pooling is a classic idea to learn representations associated with graph and can largely preserve the graph structure \cite{DBLP:conf/nips/DuvenaudMABHAA15,DBLP:conf/aaai/ChenPHS18}.
There are a few works that leverage the idea for NLP tasks. \cite{gao2019learning} adopts graph pooling to aggregate local features for global text representation, via an architecture without the unpooling operation, which differs from our work. \citet{nguyen2018graph} has also explored the idea of pooling with GCN, but their pooling is conducted on features instead of the input graph. To the best of our knowledge, we are the first to apply the idea of graph pooling to relation extraction task.

\paragraph{Pooling-unpooling Mechanism.}
The pooling-unpooling mechanism is widely-used for pixel-wise representation learning \cite{badrinarayanan2017segnet,chen2018encoder}, which use downsampling and upsampling operation to aggregate information from different resolution. The flagship work of such paradigm is U-Net \cite{DBLP:conf/miccai/RonnebergerFB15}, which demonstrates the effectiveness in the image segmentation. It is worth mentioning that \cite{DBLP:conf/icml/GaoJ19,DBLP:conf/ijcai/Hu0WWT19} also adopt such paradigm for graph node and graph classification task, and shares similarity with our work from the architectural perspective. However, the main contribution of them is the specially-designed pooling and unpooling operations or the application toward semi-supervised setting. In contrast, our work focuses on studying the effectiveness of such framework in DRE to help alleviate the efforts for GNNs to capture long-term dependencies when given long context. Besides, we propose a domain-specific designed pooling method and show competitive results.
\section{Conclusion and Future Work}
In this work, we explore the effectiveness of applying graph pooling-unpooling mechanism for learning better mention representations for DRE tasks.
Such paradigm help the used graph neural networks capture long dependencies given large input document graphs. Besides, we introduce a new graph pooling strategy that is tailored for NLP tasks. 

For the future work, we plan to explore the possibility of applying the pooling-unpooling mechanism to other document-level NLP tasks. We also plan to propose differentiable, feature-selection free pooling methods that consider edge types to better handle NLP tasks.
\bibliography{aaai22}

\begin{thebibliography}{49}
\providecommand{\natexlab}[1]{#1}

\bibitem[{Al{-}Zaidy and Giles(2017)}]{DBLP:conf/doceng/Al-ZaidyG17}
Al{-}Zaidy, R.~A.; and Giles, C.~L. 2017.
\newblock Automatic Knowledge Base Construction from Scholarly Documents.
\newblock In \emph{Proceedings of the 2017 {ACM} Symposium on Document
  Engineering (DocEng)}.

\bibitem[{Badrinarayanan, Kendall, and
  Cipolla(2017)}]{badrinarayanan2017segnet}
Badrinarayanan, V.; Kendall, A.; and Cipolla, R. 2017.
\newblock Segnet: A deep convolutional encoder-decoder architecture for image
  segmentation.
\newblock \emph{IEEE transactions on pattern analysis and machine
  intelligence}.

\bibitem[{Beltagy, Lo, and Cohan(2019)}]{DBLP:conf/emnlp/BeltagyLC19}
Beltagy, I.; Lo, K.; and Cohan, A. 2019.
\newblock SciBERT: {A} Pretrained Language Model for Scientific Text.
\newblock In \emph{Proceedings of the 2019 Conference on Empirical Methods in
  Natural Language Processing and the 9th International Joint Conference on
  Natural Language Processing, {EMNLP-IJCNLP} 2019, Hong Kong, China, November
  3-7, 2019}.

\bibitem[{Chen et~al.(2018{\natexlab{a}})Chen, Perozzi, Hu, and
  Skiena}]{DBLP:conf/aaai/ChenPHS18}
Chen, H.; Perozzi, B.; Hu, Y.; and Skiena, S. 2018{\natexlab{a}}.
\newblock {HARP:} Hierarchical Representation Learning for Networks.
\newblock In \emph{Proceedings of the Thirty-Second {AAAI} Conference on
  Artificial Intelligence, (AAAI)}.

\bibitem[{Chen et~al.(2018{\natexlab{b}})Chen, Zhu, Papandreou, Schroff, and
  Adam}]{chen2018encoder}
Chen, L.-C.; Zhu, Y.; Papandreou, G.; Schroff, F.; and Adam, H.
  2018{\natexlab{b}}.
\newblock Encoder-decoder with atrous separable convolution for semantic image
  segmentation.
\newblock In \emph{Proceedings of the European conference on computer vision
  (ECCV)}.

\bibitem[{Chiu et~al.(2016)Chiu, Crichton, Korhonen, and
  Pyysalo}]{DBLP:conf/bionlp/ChiuCKP16}
Chiu, B.; Crichton, G. K.~O.; Korhonen, A.; and Pyysalo, S. 2016.
\newblock How to Train good Word Embeddings for Biomedical {NLP}.
\newblock In \emph{Proceedings of the 15th Workshop on Biomedical Natural
  Language Processing, BioNLP@ACL}.

\bibitem[{Christopoulou, Miwa, and
  Ananiadou(2019)}]{DBLP:conf/emnlp/ChristopoulouMA19}
Christopoulou, F.; Miwa, M.; and Ananiadou, S. 2019.
\newblock Connecting the Dots: Document-level Neural Relation Extraction with
  Edge-oriented Graphs.
\newblock In \emph{Proceedings of the 2019 Conference on Empirical Methods in
  Natural Language Processing and the 9th International Joint Conference on
  Natural Language Processing, {EMNLP-IJCNLP} 2019, Hong Kong, China, November
  3-7, 2019}.

\bibitem[{Devlin et~al.(2019)Devlin, Chang, Lee, and Toutanova}]{BERT}
Devlin, J.; Chang, M.; Lee, K.; and Toutanova, K. 2019.
\newblock {BERT:} Pre-training of Deep Bidirectional Transformers for Language
  Understanding.
\newblock In \emph{Proceedings of the 2019 Conference of the North American
  Chapter of the Association for Computational Linguistics: Human Language
  Technologies (NAACL-HLT)}.

\bibitem[{Dong et~al.(2014)Dong, Gabrilovich, Heitz, Horn, Lao, Murphy,
  Strohmann, Sun, and Zhang}]{dong2014knowledge}
Dong, X.; Gabrilovich, E.; Heitz, G.; Horn, W.; Lao, N.; Murphy, K.; Strohmann,
  T.; Sun, S.; and Zhang, W. 2014.
\newblock Knowledge vault: A web-scale approach to probabilistic knowledge
  fusion.
\newblock In \emph{Proceedings of the 20th ACM SIGKDD international conference
  on Knowledge discovery and data mining}.

\bibitem[{Duvenaud et~al.(2015)Duvenaud, Maclaurin, Aguilera{-}Iparraguirre,
  G{\'{o}}mez{-}Bombarelli, Hirzel, Aspuru{-}Guzik, and
  Adams}]{DBLP:conf/nips/DuvenaudMABHAA15}
Duvenaud, D.; Maclaurin, D.; Aguilera{-}Iparraguirre, J.;
  G{\'{o}}mez{-}Bombarelli, R.; Hirzel, T.; Aspuru{-}Guzik, A.; and Adams,
  R.~P. 2015.
\newblock Convolutional Networks on Graphs for Learning Molecular Fingerprints.
\newblock In \emph{Advances in Neural Information Processing Systems 28: Annual
  Conference on Neural Information Processing Systems 2015, December 7-12,
  2015, Montreal, Quebec, Canada}.

\bibitem[{Gao, Chen, and Ji(2019)}]{gao2019learning}
Gao, H.; Chen, Y.; and Ji, S. 2019.
\newblock Learning graph pooling and hybrid convolutional operations for text
  representations.
\newblock In \emph{The World Wide Web Conference}.

\bibitem[{Gao and Ji(2019)}]{DBLP:conf/icml/GaoJ19}
Gao, H.; and Ji, S. 2019.
\newblock Graph U-Nets.
\newblock In \emph{Proceedings of the 36th International Conference on Machine
  Learning (ICML)}.

\bibitem[{Gu et~al.(2017)Gu, Sun, Qian, and Zhou}]{DBLP:journals/biodb/GuSQZ17}
Gu, J.; Sun, F.; Qian, L.; and Zhou, G. 2017.
\newblock Chemical-induced disease relation extraction via convolutional neural
  network.
\newblock \emph{Database J. Biol. Databases Curation}.

\bibitem[{Guo et~al.(2020)Guo, Nan, Lu, and Cohen}]{DBLP:conf/ijcai/GuoN0C20}
Guo, Z.; Nan, G.; Lu, W.; and Cohen, S.~B. 2020.
\newblock Learning Latent Forests for Medical Relation Extraction.
\newblock In \emph{Proceedings of the Twenty-Ninth International Joint
  Conference on Artificial Intelligence (IJCAI)}.

\bibitem[{Guo, Zhang, and Lu(2019)}]{DBLP:conf/acl/GuoZL19}
Guo, Z.; Zhang, Y.; and Lu, W. 2019.
\newblock Attention Guided Graph Convolutional Networks for Relation
  Extraction.
\newblock In \emph{Proceedings of the 57th Conference of the Association for
  Computational Linguistics (ACL)}.

\bibitem[{Gupta et~al.(2019)Gupta, Rajaram, Sch{\"{u}}tze, and
  Runkler}]{DBLP:conf/aaai/GuptaRSR19}
Gupta, P.; Rajaram, S.; Sch{\"{u}}tze, H.; and Runkler, T.~A. 2019.
\newblock Neural Relation Extraction within and across Sentence Boundaries.
\newblock In \emph{The Thirty-Third {AAAI} Conference on Artificial
  Intelligence (AAAI)}.

\bibitem[{He et~al.(2016)He, Zhang, Ren, and Sun}]{DBLP:conf/cvpr/HeZRS16}
He, K.; Zhang, X.; Ren, S.; and Sun, J. 2016.
\newblock Deep Residual Learning for Image Recognition.
\newblock In \emph{2016 {IEEE} Conference on Computer Vision and Pattern
  Recognition, {CVPR} 2016, Las Vegas, NV, USA, June 27-30, 2016}.

\bibitem[{Hu et~al.(2019)Hu, Zhu, Wu, Wang, and Tan}]{DBLP:conf/ijcai/Hu0WWT19}
Hu, F.; Zhu, Y.; Wu, S.; Wang, L.; and Tan, T. 2019.
\newblock Hierarchical Graph Convolutional Networks for Semi-supervised Node
  Classification.
\newblock In \emph{Proceedings of the Twenty-Eighth International Joint
  Conference on Artificial Intelligence, {IJCAI} 2019, Macao, China, August
  10-16, 2019}.

\bibitem[{Jia, Wong, and Poon(2019)}]{DBLP:conf/naacl/JiaWP19}
Jia, R.; Wong, C.; and Poon, H. 2019.
\newblock Document-Level N-ary Relation Extraction with Multiscale
  Representation Learning.
\newblock In \emph{Proceedings of the 2019 Conference of the North American
  Chapter of the Association for Computational Linguistics: Human Language
  Technologies (NAACL-HLT)}.

\bibitem[{Kipf and Welling(2017)}]{DBLP:conf/iclr/KipfW17}
Kipf, T.~N.; and Welling, M. 2017.
\newblock Semi-Supervised Classification with Graph Convolutional Networks.
\newblock In \emph{5th International Conference on Learning Representations
  (ICLR)}.

\bibitem[{Lee, Lee, and Kang(2019)}]{DBLP:conf/icml/LeeLK19}
Lee, J.; Lee, I.; and Kang, J. 2019.
\newblock Self-Attention Graph Pooling.
\newblock In \emph{Proceedings of the 36th International Conference on Machine
  Learning, {ICML} 2019, 9-15 June 2019, Long Beach, California, {USA}}.

\bibitem[{Li et~al.(2016)Li, Sun, Johnson, Sciaky, Wei, Leaman, Davis,
  Mattingly, Wiegers, and Lu}]{CDR}
Li, J.; Sun, Y.; Johnson, R.~J.; Sciaky, D.; Wei, C.; Leaman, R.; Davis, A.~P.;
  Mattingly, C.~J.; Wiegers, T.~C.; and Lu, Z. 2016.
\newblock BioCreative {V} {CDR} task corpus: a resource for chemical disease
  relation extraction.
\newblock \emph{Database J. Biol. Databases Curation}.

\bibitem[{Li, Han, and Wu(2018)}]{DBLP:conf/aaai/LiHW18}
Li, Q.; Han, Z.; and Wu, X. 2018.
\newblock Deeper Insights Into Graph Convolutional Networks for Semi-Supervised
  Learning.
\newblock In \emph{Proceedings of the Thirty-Second Conference on Artificial
  Intelligence (AAAI)}.

\bibitem[{Liang, Gurukar, and Parthasarathy(2021)}]{DBLP:conf/icwsm/LiangG021}
Liang, J.; Gurukar, S.; and Parthasarathy, S. 2021.
\newblock {MILE:} {A} Multi-Level Framework for Scalable Graph Embedding.
\newblock In \emph{Proceedings of the Fifteenth International {AAAI} Conference
  on Web and Social Media (ICWSM)}.

\bibitem[{Loshchilov and Hutter(2019)}]{DBLP:conf/iclr/LoshchilovH19}
Loshchilov, I.; and Hutter, F. 2019.
\newblock Decoupled Weight Decay Regularization.
\newblock In \emph{7th International Conference on Learning Representations
  (ICLR)}.

\bibitem[{Luo et~al.(2016)Luo, Li, Urtasun, and
  Zemel}]{DBLP:conf/nips/LuoLUZ16}
Luo, W.; Li, Y.; Urtasun, R.; and Zemel, R.~S. 2016.
\newblock Understanding the Effective Receptive Field in Deep Convolutional
  Neural Networks.
\newblock In \emph{Advances in Neural Information Processing Systems 29: Annual
  Conference on Neural Information Processing Systems 2016 (NeurIPS)}.

\bibitem[{Manning et~al.(2014)Manning, Surdeanu, Bauer, Finkel, Bethard, and
  McClosky}]{DBLP:conf/acl/ManningSBFBM14}
Manning, C.~D.; Surdeanu, M.; Bauer, J.; Finkel, J.~R.; Bethard, S.; and
  McClosky, D. 2014.
\newblock The Stanford CoreNLP Natural Language Processing Toolkit.
\newblock In \emph{Proceedings of the 52nd Annual Meeting of the Association
  for Computational Linguistics, {ACL} 2014, June 22-27, 2014, Baltimore, MD,
  USA, System Demonstrations}.

\bibitem[{Mintz et~al.(2009)Mintz, Bills, Snow, and
  Jurafsky}]{DBLP:conf/acl/MintzBSJ09}
Mintz, M.; Bills, S.; Snow, R.; and Jurafsky, D. 2009.
\newblock Distant supervision for relation extraction without labeled data.
\newblock In \emph{{ACL} 2009, Proceedings of the 47th Annual Meeting of the
  Association for Computational Linguistics and the 4th International Joint
  Conference on Natural Language Processing of the AFNLP, 2-7 August 2009,
  Singapore}.

\bibitem[{Miwa and Bansal(2016)}]{DBLP:conf/acl/MiwaB16}
Miwa, M.; and Bansal, M. 2016.
\newblock End-to-End Relation Extraction using LSTMs on Sequences and Tree
  Structures.
\newblock In \emph{Proceedings of the 54th Annual Meeting of the Association
  for Computational Linguistics (ACL)}.

\bibitem[{Nan et~al.(2020)Nan, Guo, Sekulic, and Lu}]{DBLP:conf/acl/NanGSL20}
Nan, G.; Guo, Z.; Sekulic, I.; and Lu, W. 2020.
\newblock Reasoning with Latent Structure Refinement for Document-Level
  Relation Extraction.
\newblock In \emph{Proceedings of the 58th Annual Meeting of the Association
  for Computational Linguistics (ACL)}.

\bibitem[{Nguyen and Grishman(2018)}]{nguyen2018graph}
Nguyen, T.~H.; and Grishman, R. 2018.
\newblock Graph convolutional networks with argument-aware pooling for event
  detection.
\newblock In \emph{Thirty-second AAAI conference on artificial intelligence}.

\bibitem[{Nivre et~al.(2020)Nivre, de~Marneffe, Ginter, Hajic, Manning,
  Pyysalo, Schuster, Tyers, and Zeman}]{DBLP:conf/lrec/NivreMGHMPSTZ20}
Nivre, J.; de~Marneffe, M.; Ginter, F.; Hajic, J.; Manning, C.~D.; Pyysalo, S.;
  Schuster, S.; Tyers, F.~M.; and Zeman, D. 2020.
\newblock Universal Dependencies v2: An Evergrowing Multilingual Treebank
  Collection.
\newblock In \emph{Proceedings of The 12th Language Resources and Evaluation
  Conference (LREC)}.

\bibitem[{Paszke et~al.(2019)Paszke, Gross, Massa, Lerer, Bradbury, Chanan,
  Killeen, Lin, Gimelshein, Antiga, Desmaison, K{\"{o}}pf, Yang, DeVito,
  Raison, Tejani, Chilamkurthy, Steiner, Fang, Bai, and
  Chintala}]{DBLP:conf/nips/PaszkeGMLBCKLGA19}
Paszke, A.; Gross, S.; Massa, F.; Lerer, A.; Bradbury, J.; Chanan, G.; Killeen,
  T.; Lin, Z.; Gimelshein, N.; Antiga, L.; Desmaison, A.; K{\"{o}}pf, A.; Yang,
  E.; DeVito, Z.; Raison, M.; Tejani, A.; Chilamkurthy, S.; Steiner, B.; Fang,
  L.; Bai, J.; and Chintala, S. 2019.
\newblock PyTorch: An Imperative Style, High-Performance Deep Learning Library.
\newblock In \emph{Advances in Neural Information Processing Systems 32: Annual
  Conference on Neural Information Processing Systems 2019 (Neurips)}.

\bibitem[{Peng et~al.(2017)Peng, Poon, Quirk, Toutanova, and Yih}]{nary}
Peng, N.; Poon, H.; Quirk, C.; Toutanova, K.; and Yih, W. 2017.
\newblock Cross-Sentence N-ary Relation Extraction with Graph LSTMs.
\newblock \emph{Trans. Assoc. Comput. Linguistics}.

\bibitem[{Pennington, Socher, and
  Manning(2014)}]{DBLP:conf/emnlp/PenningtonSM14}
Pennington, J.; Socher, R.; and Manning, C.~D. 2014.
\newblock Glove: Global Vectors for Word Representation.
\newblock In \emph{Proceedings of the 2014 Conference on Empirical Methods in
  Natural Language Processing (EMNLP)}.

\bibitem[{Quirk and Poon(2017)}]{DBLP:conf/eacl/QuirkP17}
Quirk, C.; and Poon, H. 2017.
\newblock Distant Supervision for Relation Extraction beyond the Sentence
  Boundary.
\newblock In \emph{Proceedings of the 15th Conference of the European Chapter
  of the Association for Computational Linguistics, {EACL} 2017, Valencia,
  Spain, April 3-7, 2017, Volume 1: Long Papers}.

\bibitem[{Ronneberger, Fischer, and
  Brox(2015)}]{DBLP:conf/miccai/RonnebergerFB15}
Ronneberger, O.; Fischer, P.; and Brox, T. 2015.
\newblock U-Net: Convolutional Networks for Biomedical Image Segmentation.
\newblock In \emph{Medical Image Computing and Computer-Assisted Intervention -
  {MICCAI} 2015 - 18th International Conference Munich, Germany, October 5 - 9,
  2015, Proceedings, Part {III}}.

\bibitem[{Sahu et~al.(2019)Sahu, Christopoulou, Miwa, and
  Ananiadou}]{DBLP:conf/acl/SahuCMA19}
Sahu, S.~K.; Christopoulou, F.; Miwa, M.; and Ananiadou, S. 2019.
\newblock Inter-sentence Relation Extraction with Document-level Graph
  Convolutional Neural Network.
\newblock In \emph{Proceedings of the 57th Conference of the Association for
  Computational Linguistics, {ACL} 2019, Florence, Italy, July 28- August 2,
  2019, Volume 1: Long Papers}.

\bibitem[{Song et~al.(2018)Song, Zhang, Wang, and
  Gildea}]{DBLP:conf/emnlp/SongZWG18}
Song, L.; Zhang, Y.; Wang, Z.; and Gildea, D. 2018.
\newblock N-ary Relation Extraction using Graph-State {LSTM}.
\newblock In \emph{Proceedings of the 2018 Conference on Empirical Methods in
  Natural Language Processing (EMNLP)}.

\bibitem[{Tran, Nguyen, and Nguyen(2020)}]{DBLP:conf/emnlp/TranNN20}
Tran, H.~M.; Nguyen, T.~M.; and Nguyen, T.~H. 2020.
\newblock The Dots Have Their Values: Exploiting the Node-Edge Connections in
  Graph-based Neural Models for Document-level Relation Extraction.
\newblock In \emph{Proceedings of the 2020 Conference on Empirical Methods in
  Natural Language Processing: Findings (EMNLP)}.

\bibitem[{Tsuruoka et~al.(2005)Tsuruoka, Tateishi, Kim, Ohta, McNaught,
  Ananiadou, and Tsujii}]{tsuruoka2005developing}
Tsuruoka, Y.; Tateishi, Y.; Kim, J.-D.; Ohta, T.; McNaught, J.; Ananiadou, S.;
  and Tsujii, J. 2005.
\newblock Developing a robust part-of-speech tagger for biomedical text.
\newblock In \emph{Panhellenic Conference on Informatics}, 382--392. Springer.

\bibitem[{Wolf et~al.(2019)Wolf, Debut, Sanh, Chaumond, Delangue, Moi, Cistac,
  Rault, Louf, Funtowicz, and Brew}]{DBLP:journals/corr/abs-1910-03771}
Wolf, T.; Debut, L.; Sanh, V.; Chaumond, J.; Delangue, C.; Moi, A.; Cistac, P.;
  Rault, T.; Louf, R.; Funtowicz, M.; and Brew, J. 2019.
\newblock HuggingFace's Transformers: State-of-the-art Natural Language
  Processing.
\newblock \emph{arXiv preprint arXiv:1910.03771}.

\bibitem[{Xu et~al.(2021)Xu, Wang, Lyu, Zhu, and Mao}]{DBLP:conf/aaai/XuWLZM21}
Xu, B.; Wang, Q.; Lyu, Y.; Zhu, Y.; and Mao, Z. 2021.
\newblock Entity Structure Within and Throughout: Modeling Mention Dependencies
  for Document-Level Relation Extraction.
\newblock In \emph{Thirty-Fifth {AAAI} Conference on Artificial Intelligence
  (AAAI)}.

\bibitem[{Xu et~al.(2015)Xu, Feng, Huang, and Zhao}]{DBLP:conf/emnlp/XuFHZ15}
Xu, K.; Feng, Y.; Huang, S.; and Zhao, D. 2015.
\newblock Semantic Relation Classification via Convolutional Neural Networks
  with Simple Negative Sampling.
\newblock In \emph{Proceedings of the 2015 Conference on Empirical Methods in
  Natural Language Processing (EMNLP)}.

\bibitem[{Xu, Chen, and Zhao(2021)}]{DBLP:conf/aaai/XuCZ21}
Xu, W.; Chen, K.; and Zhao, T. 2021.
\newblock Document-Level Relation Extraction with Reconstruction.
\newblock In \emph{Thirty-Fifth {AAAI} Conference on Artificial Intelligence
  (AAAI)}.

\bibitem[{Ying et~al.(2018)Ying, You, Morris, Ren, Hamilton, and
  Leskovec}]{DBLP:conf/nips/YingY0RHL18}
Ying, Z.; You, J.; Morris, C.; Ren, X.; Hamilton, W.~L.; and Leskovec, J. 2018.
\newblock Hierarchical Graph Representation Learning with Differentiable
  Pooling.
\newblock In \emph{Advances in Neural Information Processing Systems 31: Annual
  Conference on Neural Information Processing Systems 2018 (NeurIPS)}.

\bibitem[{Zeng et~al.(2020)Zeng, Xu, Chang, and Li}]{DBLP:conf/emnlp/ZengXCL20}
Zeng, S.; Xu, R.; Chang, B.; and Li, L. 2020.
\newblock Double Graph Based Reasoning for Document-level Relation Extraction.
\newblock In \emph{Proceedings of the 2020 Conference on Empirical Methods in
  Natural Language Processing (EMNLP)}.

\bibitem[{Zhang, Qi, and Manning(2018)}]{DBLP:conf/emnlp/Zhang0M18}
Zhang, Y.; Qi, P.; and Manning, C.~D. 2018.
\newblock Graph Convolution over Pruned Dependency Trees Improves Relation
  Extraction.
\newblock In \emph{Proceedings of the 2018 Conference on Empirical Methods in
  Natural Language Processing (EMNLP)}.

\bibitem[{Zhou et~al.(2021)Zhou, Huang, Ma, and
  Huang}]{DBLP:conf/aaai/Zhou0M021}
Zhou, W.; Huang, K.; Ma, T.; and Huang, J. 2021.
\newblock Document-Level Relation Extraction with Adaptive Thresholding and
  Localized Context Pooling.
\newblock In \emph{Thirty-Fifth {AAAI} Conference on Artificial Intelligence
  (AAAI)}.

\end{thebibliography}

\clearpage

\appendix
\section{$n$-ary results under Entity Identity}
\label{appendix:entity_id_nary}
Table~\ref{tab:identity} presents the models' performance on the $n$-ary dataset under Entity Identity setup. We can observe that our model can achieve extremely well performance because it can simply memorized the entity name information to make predictions given the $n$-ary dataset is curated via distant supervision.

\begin{table*}[t!]
\centering
\small
\begin{tabular}{l|l|cccc}\toprule
\multicolumn{2}{l|}{\multirow{2}{*}{\textbf{Model}}}                                                               & \multicolumn{2}{c}{\textbf{Detection}} & \multicolumn{2}{c}{\textbf{Classification}} \\ \cline{3-6} 
\multicolumn{2}{l|}{}                                                                                     & Ternary           & Binary          & Ternary             & Binary            \\ \hline
\multicolumn{2}{l|}{GS GLSTM \cite{DBLP:conf/emnlp/SongZWG18}}                                                                           & 83.2           & 83.6          & 71.7               & 71.7               \\ 
\multicolumn{2}{l|}{GCN (Full Tree) \cite{DBLP:conf/emnlp/Zhang0M18}}                                                                           & 84.8           & 83.6          & 77.5               & 74.3              \\ 
\multicolumn{2}{l|}{AGGCN \cite{DBLP:conf/acl/GuoZL19}}                                                                                & 87.0           & 85.6          & 79.7           & 77.4            \\ \hline
\multicolumn{2}{l|}{Pooling-Unpooling (HM)$_{\text{GloVe}}$}                                                                                & \textbf{91.5}           & \textbf{94.0}          & \textbf{90.8}        & \textbf{94.7}         \\
\multicolumn{2}{l|}{Pooling-Unpooling (CM)$_{\text{GloVe}}$}                                                                                & 91.4      & 93.9         & \textbf{90.8}        & 93.6        \\

\bottomrule
\end{tabular}
\caption{Results in average accuracy (\%) of five-fold cross validation on the $n$-ary dataset for 4 sub-tasks when we use Entity Identity settings.}
\label{tab:identity}
\end{table*}

\section{Dataset Details}
\label{data_detail}
\subsection{$n$-ary dataset}
The statistics of the $n$-ary dataset is presented in Table~\ref{sta_nary} and Table~\ref{special_stat_nary}.

\begin{table}[h!]
\small
\centering
\begin{tabular}{l|c c }\toprule
Folds & Binary & Ternary \\ \hline
fold\#0 & 1256 & 1474 \\  
fold\#1 & 1180 & 1432 \\
fold\#2 & 1234 & 1252 \\
fold\#3 & 1206 & 1531 \\
fold\#4 & 1211 & 1298 \\
\bottomrule
\end{tabular}
\caption{The detailed distribution of data instances over the 5-fold on the $n$-ary dataset.} 
\label{sta_nary}
\end{table}
\begin{table}[h!]
\small
\centering
\begin{tabular}{l|c |c | c}\toprule
Data & Avg. Token & Avg. Sent. & Cross \\ \hline
Ternary & 73.0 & 2.0 & 70.1\% \\
Binary & 61.0 & 1.8 & 55.2\% \\
\bottomrule
\end{tabular}
\caption{The $n$-ary dataset statistics. ``Avg. Token'' and ``Avg. Sent.'' are the average number of tokens and the average sentence length per instance. ``Cross'' means the percentage of instance that contains multiple sentences.} 
\label{special_stat_nary}
\end{table}

\subsection{CDR dataset}
We adapt our data-preprocessing from the source released by \citet{DBLP:conf/emnlp/ChristopoulouMA19}\footnote{https://github.com/fenchri/edge-oriented-graph}. The statistics is presented in Table~\ref{tab:cdr_stat}. We also follow \citet{DBLP:journals/biodb/GuSQZ17,DBLP:conf/emnlp/ChristopoulouMA19} to ignore non-related pairs that correspond to general concepts (MeSH vocabulary hypernym filtering). More details can refer to \cite{DBLP:journals/biodb/GuSQZ17}.

\begin{table}[h!]
\centering
\begin{tabular}{ll|rrr}\toprule
        &                                & Train & Dev  & Test \\ \hline
\multicolumn{2}{l|}{Documents}           & 500   & 500  & 500  \\
\multicolumn{2}{l|}{Positive pairs}      & 1038  & 1012 & 1066 \\
\multicolumn{2}{r|}{Intra}               & 754   & 766  & 747  \\
        & \multicolumn{1}{r|}{Inter}     & 284   & 246  & 319  \\
\multicolumn{2}{l|}{Negative pairs}      & 4202  & 4075 & 4138 \\
\multicolumn{2}{l|}{Entities}            &       &      &      \\
\multicolumn{2}{r|}{Chemical}            & 1467  & 1507 & 1434 \\
\multicolumn{2}{r|}{Disease}             & 1965  & 1864 & 1988 \\
\multicolumn{2}{l|}{Mentions}            &       &      &      \\
\multicolumn{2}{r|}{Chemical}            & 5162  & 5307 & 5370 \\
\multicolumn{2}{r|}{Disease}             & 4252  & 4328 & 4430 \\
\multicolumn{2}{l|}{Avg sent. len./doc.} & 25.6  & 25.4 & 25.7 \\
\multicolumn{2}{l|}{Avg sents./doc.}     & 9.2   & 9.3  & 9.7 \\                
\bottomrule
\end{tabular}
\caption{CDR dataset statistics}
\label{tab:cdr_stat}
\end{table}

\section{Implementation Details}
\label{hyper_detail}
Our models are developed using PyTorch~\cite{DBLP:conf/nips/PaszkeGMLBCKLGA19} and SGD optimizer in the experiments with static embeddings. As for other experiments with contextualized word representations, we use AdamW~\cite{DBLP:conf/iclr/LoshchilovH19} optimizer.
Dataset-specific implementation details are provided as follows.
\subsection{$n$-ary dataset}
\subsubsection{Model architecture}
For models that use GloVE embeddings, the word representations are with 300 dimensions. As for models with contextualized word representations, the dimension of the word vectors depends on the type of contextualized embedding. The pretrained models are get from Huggingface library~\cite{DBLP:journals/corr/abs-1910-03771}. We further concatenate the word representations with 30-dimensional POS tag embedding before feeding into the pooling-unpooling layers. All these vectors will be updated during training. We use one layer Bi-LSTM with 330 hidden dimensions for each GCN layers. Dropout layers are used to prevent overfitting and are set to 0.5. 

\subsubsection{Hyper-parameters}
The learning rate of SGD optimizer is initialized at 0.1 with 0.95 decay after 15 epochs. When using the AdamW optimizer, the learning rate is set with 0.00005 for contextualized word embedding layers and 0.001 for other layers. We report our best hyperparameter -- the number of GCN layers in GCN block (``Sublayer'' in Table~\ref{best-hyper_nary}) and the number of pooling times ($L$ in the main article, ``Level'' in Table~\ref{best-hyper_nary}) for each different tasks in Table~\ref{best-hyper_nary}. The search range for the hyperparameter is $L=\{2,3,4\}$ and $S=\{1,2\}$ for models using contextualized embeddings and $S=\{2,3,4\}$ for models using static embeddings. The search range of the layers in GCN is set in the range $\{2,4,6,8,10,12\}$.

\begin{table}[t!]
\small
\centering
\resizebox{\columnwidth}{!}{
\begin{tabular}{l| c c| c c }
\toprule
 & \multicolumn{2}{c|}{{Detection}} & \multicolumn{2}{c}{Classification} \\
& Ternary & Binary & Ternary & Binary \\ \hline
Pooling-Unpooling (CM)$_{\text{SciBERT}}$ &L3S2 & L2S1 & L3S2 & L2S2 \\
Pooling-Unpooling (HM)$_{\text{SciBERT}}$ &L3S1 & L3S2 & L2S1 & L2S1 \\
GCN$_{\text{SciBERT}}$ & 4 & 10& 10& 4 \\
Pooling-Unpooling (CM)$_{\text{GloVe}}$& L3S2 & L3S3 & L3S2 & L3S2 \\
Pooling-Unpooling (HM)$_{\text{GloVe}}$&L4S2 & L4S2 & L4S2 & L4S2 \\
\bottomrule
\end{tabular}}
\centering
\caption{The best Sublayer ($S$) and Level ($L$) for models in our results for $n$-ary dataset. The total number of GCN layers can be calculated by $(2L-1)\times S$. For models without graph pooling, we list the total number of GCN layers we use.} 
\label{best-hyper_nary}
\end{table}

\subsection{CDR dataset}
\subsubsection{Model architecture}
We follow \citet{DBLP:conf/emnlp/ChristopoulouMA19} to use the PubMed pre-trained word embedding in the experiment with static embeddings and we fix it during training. The setup of the GCN layers, and dropout layer are the similar to the setting in the $n$-ary dataset. For contextualized word representations, we use packages from Huggingface library.

\subsubsection{Hyper-parameters}
We reuse the hyperparameter symbol of $L$ and $S$ as them in $n$-ary dataset. Then, we list the best hyper-parameter in Table~\ref{best-hyper_cdr}. The search range for the hyperparameter is $L=\{2,3,4\}$ and $S=\{1,2\}$ for models using contextualized embeddings and $S=\{2,3,4\}$ for models using static embeddings. The search range of the layers in GCN experiments is the same as it in the $n$-ary dataset.

\begin{table}[t!]
\small
\centering
\begin{tabular}{l| c c }
\toprule
& Layers & Epoch \\ \hline
Pooling-Unpooling (CM)$_{\text{BERT Base}}$& L3S2 & 20 \\ 
Pooling-Unpooling (HM)$_{\text{BERT Base}}$& L4S2 & 21 \\
GCN$_{\text{BERT Base}}$& 4 & 13 \\ 
Pooling-Unpooling (CM)$_{\text{BERT Large}}$& L3S1 & 23 \\ 
Pooling-Unpooling (HM)$_{\text{BERT Large}}$& L3S2 & 24 \\
GCN$_{\text{BERT Large}}$& 10 & 16 \\ 
Pooling-Unpooling (CM)$_{\text{SciBERT}}$& L4S2 & 16 \\ 
Pooling-Unpooling (HM)$_{\text{SciBERT}}$& L4S2 & 21 \\
GCN$_{\text{SciBERT}}$& 10 & 16 \\ 
Pooling-Unpooling (HM)$_{\text{Static}}$& L3S4 & 55 \\ 
Pooling-Unpooling (CM)$_{\text{Static}}$& L4S4 & 74 \\ 
Pooling-Unpooling (HM)$_{\text{Static}}$& L3S4 & 55 \\ 
\bottomrule
\end{tabular}
\centering
\caption{The best Sublayer ($S$) and Level ($L$) for each model in our main result. The total number of GCN layers can be calculated by $(2L-1)\times S$. For models without graph pooling, we list the total number of GCN layers we use. We also report the epoch when each of our models gets its optimal development set performance.} 
\label{best-hyper_cdr}
\end{table}

\section{Matching Matrices}
\label{appendix:matching_matrix}
\begin{figure}[h!]
\centering
    \includegraphics[width=0.45\textwidth]{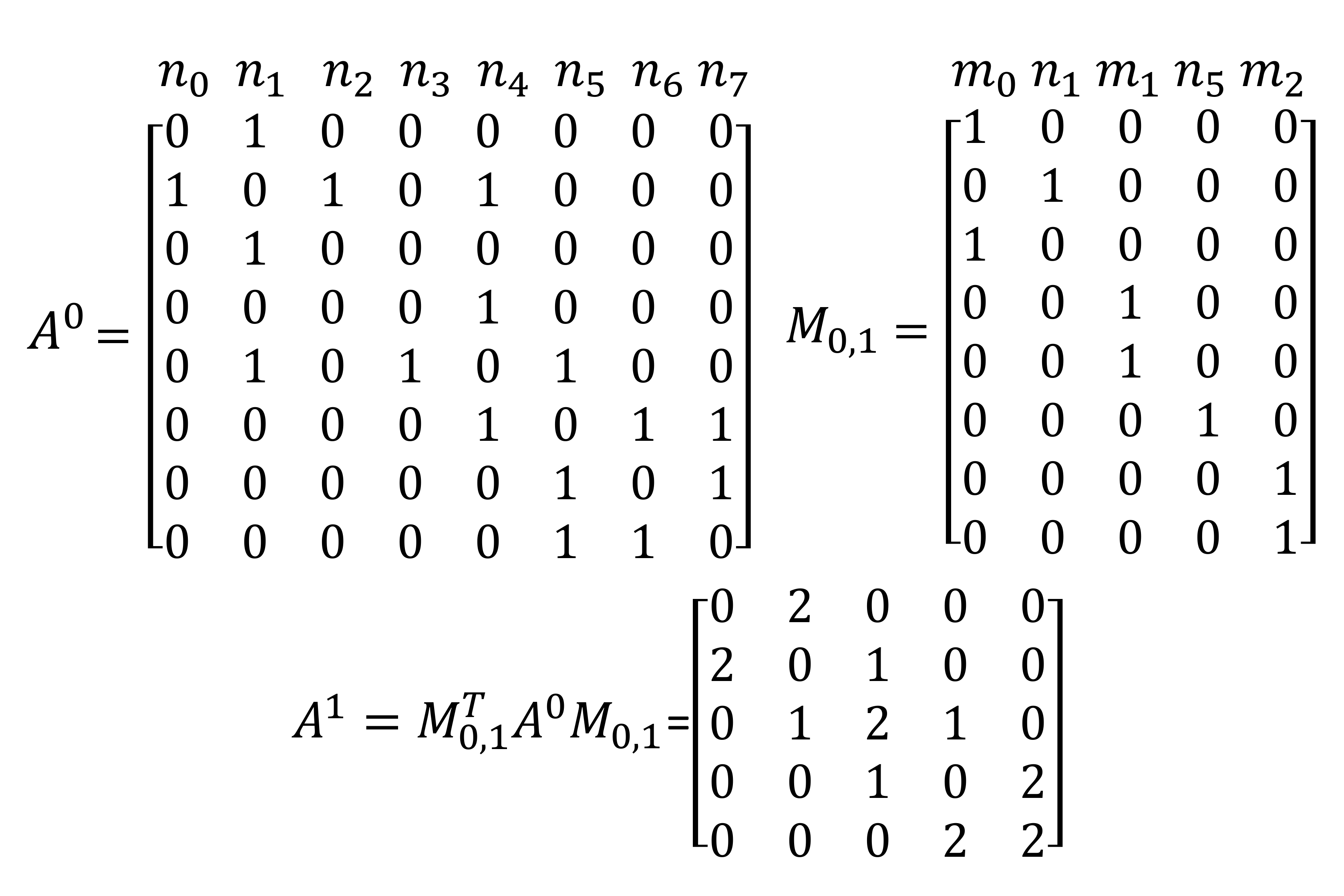}
	\caption{The matching matrix example.}
	\label{matching_matrix}
\end{figure}

In this section, we attach the corresponding matching matrices and illustrate how adjacency matrix $A^1$ can be derived from $A^0$ in Figure~\ref{matching_matrix}, given the Hybrid Matching (HM) example in Figure~\ref{fig:hybrid_matching}.

\section{Visualization of Clause Matching}
\label{appendix:visual_cm}
\begin{figure*}[t!]
\centering
    \includegraphics[width=1.0\linewidth]{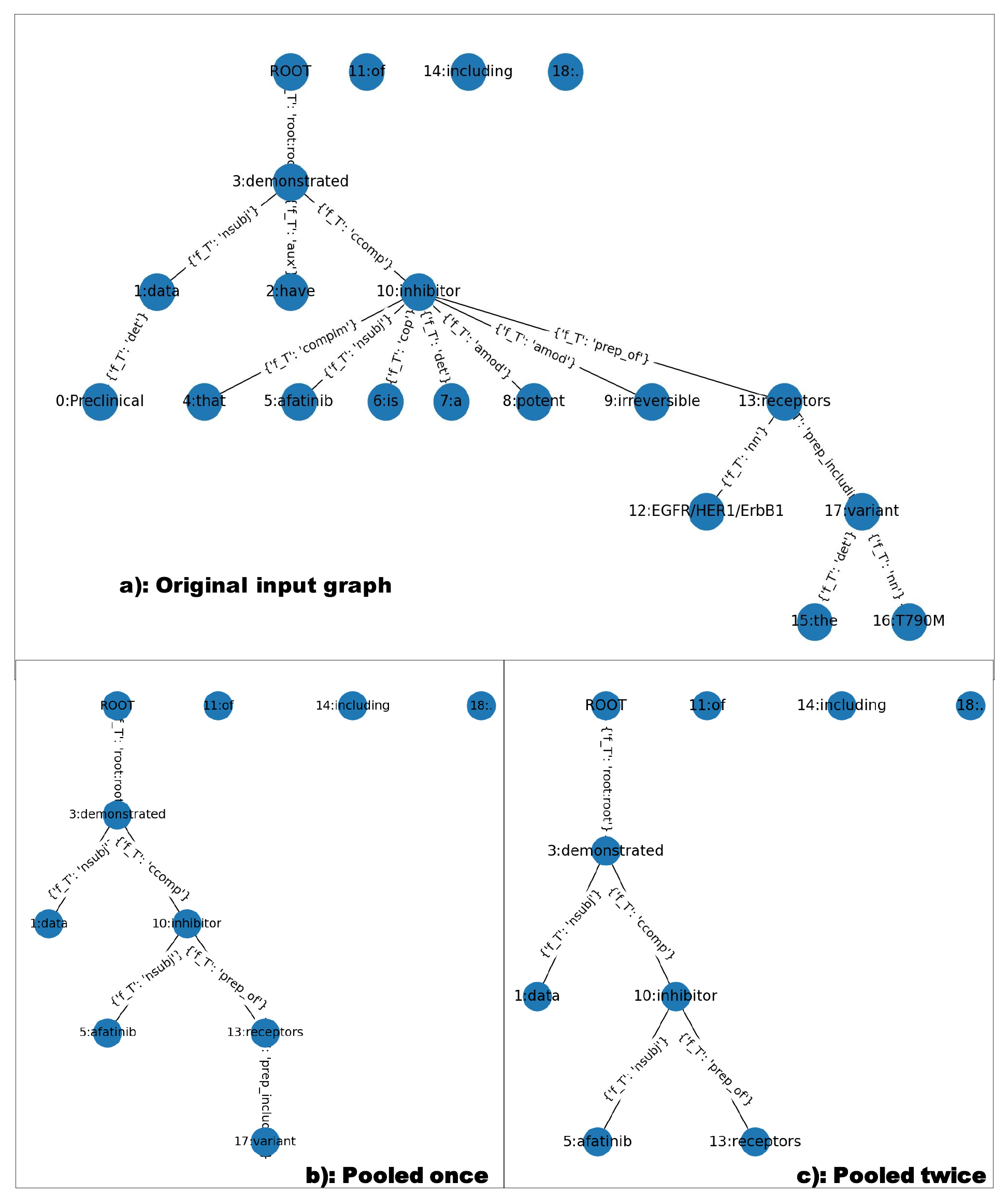}
	\caption{The visualization of Clause Matching for the input \textit{``Preclinical data have demonstrated that aftinib is a potent irreversible inhibitor of EGFR/HER1/ErbB1 receptors including the T790M variant .''}. To better visualize the result, we omit the adjacency edges in this figure. Noted that \textit{``ROOT''} does not exist in the real graph, we add the root node here in order to clearly show the tree structure.}
	\label{vis_CM_ori}
\end{figure*}

In this section, we demonstrate a real example from the $n$-ary dataset to illustrate how Clause Matching (CM) works. Figure~\ref{vis_CM_ori}a shows the original input graph in the sentence \textit{``Preclinical data have demonstrated that aftinib is a potent irreversible inhibitor of EGFR/HER1/ErbB1 receptors including the T790M variant .''}. In order to better visualize the CM process, we omit the adjacency edges in this figure. 

In Figure~\ref{vis_CM_ori}a, we can observe that nodes such as \textit{``of''}, \textit{``including''}, and \textit{``.''} are disconnected from others because several dependency arcs are dropped in the original $n$-ary dataset, as we stated in Section~\ref{N_ary_main_res}. We show the pooled graph in Figure~\ref{vis_CM_ori}b using CM algorithm given the input graph in Figure~\ref{vis_CM_ori}a. If a child node is merged with its parent forming a supernode, we use the parent node to visualize the supernode. For example, \textit{``Preclinical''} is merged with \textit{``data''}, so we only show \textit{``data''} in Figure~\ref{vis_CM_ori}b. 

After CM pooling, several non-core arguments of \textit{``inhibitor''} are merged, but we can still identify the main subject of \textit{``inihibitor''}, i.e \textit{``afatinib''}, in the graph, which demonstrates CM's ability on keeping the main component of a clause. The result of using CM pooling twice is shown in Figure~\ref{vis_CM_ori}c. As depicted in the figure, although we have pooled the graph twice and have largely cut the graph size, those isolated notes are not able to be merged. This is the reason why we hypothesize that CM does not work as well as HM in some sub-tasks of the $n$-ary dataset. 

\end{document}